\documentclass[conference]{IEEEtran}
\IEEEoverridecommandlockouts
\usepackage{cite}
\usepackage{amsmath,amssymb,amsfonts}
\usepackage{graphicx}
\usepackage{textcomp}
\usepackage{xcolor}
\usepackage{hyperref}
\usepackage{url}
\usepackage{graphicx}
\usepackage{booktabs}
\usepackage{amsfonts}
\usepackage{nicefrac}
\usepackage{microtype}
\usepackage{adjustbox}
\usepackage{subcaption}
\usepackage{listings}
\usepackage{tcolorbox}
\usepackage{amsmath} 
\usepackage{algorithm}    
\usepackage{algpseudocode}
\usepackage{hyperref}
\usepackage{url}
\usepackage{listings}
\usepackage{algpseudocode}
\usepackage{amsthm}
\theoremstyle{definition}
\newtheorem{example}{Example}
\algrenewcommand\algorithmicrequire{\textbf{Input:}}
\algrenewcommand\algorithmicensure{\textbf{Output:}}
\newtheoremstyle{plain-nobold}%
  {3pt}{3pt}{\normalfont}{}{\normalfont}{.}{0.5em}{#1~#2\ifx\relax#3\relax\else\ (#3)\fi}
\theoremstyle{plain-nobold}

\usepackage[framemethod=default]{mdframed}
\newmdenv[
  skipabove=6pt,skipbelow=6pt,
  linewidth=.6pt,roundcorner=2pt,
  innerleftmargin=6pt,innerrightmargin=6pt,innertopmargin=6pt,innerbottommargin=6pt,
  backgroundcolor=white
]{exbox}

\def\BibTeX{{\rm B\kern-.05em{\sc i\kern-.025em b}\kern-.08em
    T\kern-.1667em\lower.7ex\hbox{E}\kern-.125emX}}
\begin{document}

\title{Graph Your Way to Inspiration: Integrating Co-Author Graphs with Retrieval-Augmented Generation for Large Language Model Based Scientific Idea Generation\\
{\footnotesize \textsuperscript{}}
}

\author{\IEEEauthorblockN{ Pengzhen Xie}
\IEEEauthorblockA{\textit{School of Computing} \\
\textit{Newcastle University}\\
Newcastle upon Tyne, UK \\
p.xie4@newcastle.ac.uk}
~\\
\and
\IEEEauthorblockN{ Huizhi Liang}
\IEEEauthorblockA{\textit{School of Computing} \\
\textit{Newcastle University}\\
Newcastle upon Tyne, UK \\
Huizhi.Liang@newcastle.ac.uk}
}

\maketitle

\begin{abstract}

Large Language Models (LLMs) demonstrate potential in the field of scientific idea generation. However, the generated results often lack controllable academic context and traceable inspiration pathways.  To bridge this gap, this paper proposes a scientific idea generation system called GYWI, which combines author knowledge graphs with retrieval-augmented generation (RAG) to form an external knowledge base to provide controllable context and trace of inspiration path for LLMs to generate new scientific ideas. We first propose an author-centered knowledge graph construction method and inspiration source sampling algorithms to construct external knowledge base. Then, we propose a hybrid retrieval mechanism that is composed of both RAG and GraphRAG to retrieve content with both depth and breadth knowledge. It forms a hybrid context. Thirdly, we propose a Prompt optimization strategy incorporating reinforcement learning principles to automatically guide LLMs optimizing the results based on the hybrid context. To evaluate the proposed approaches, we constructed an evaluation dataset based on arXiv (2018-2023). This paper also develops a comprehensive evaluation method including empirical automatic assessment in multiple-choice question task, LLM-based scoring, human evaluation, and semantic space visualization analysis. The generated ideas are evaluated from the following five dimensions: novelty, feasibility, clarity, relevance, and significance. We conducted experiments on different LLMs including GPT-4o, DeepSeek-V3, Qwen3-8B, and Gemini 2.5. Experimental results show that GYWI significantly outperforms mainstream LLMs in multiple metrics such as novelty, reliability, and relevance.   

\end{abstract}


\section{Introduction}
In the scientific research process, proposing innovative research ideas constitutes a crucial stage. Traditionally, researchers have often required extensive reading of cutting-edge literature and in-depth discussions with relevant colleagues to identify new research directions. However, this process demands considerable time and effort, particularly in fields such as artificial intelligence and medicine where information evolves rapidly and broad knowledge is essential. As knowledge expands and updates at scale, human factors---including cognitive bias and limited prior knowledge---further constrain the ideation process. Consequently, generating forward-thinking and feasible ideas has become increasingly challenging.

With the rapid advancement of large language models, LLMs have demonstrated exceptional capabilities in literature reviews, knowledge extraction, and question-answering tasks; for instance, solving mathematical problems\cite{mirzadeh_gsm-symbolic_2024,xiong_self-rewarding_2025}, assisting psychological research\cite{chen_towards_2025}, and recommending literature relevant to research directions. With the further refinement of large language models over the past two years and their tighter integration with retrieval techniques, researchers have begun employing LLMs as primary tools for scientific inquiry, marking their emergence as practical instruments within scientific research\cite{lu_ai_2024,li_chain_2024}.

However, applying LLMs to scientific idea generation remains challenging. As LLM outputs primarily rely on pre-training data, they often exhibit semantic fluency yet suffer from unverifiable origins and opaque innovation pathways, while remaining susceptible to ``hallucinations'' \cite{guo_ideabench_2024,qiu_ai_2025,sun_scieval_2024}. Consequently, generated research ideas frequently lack controllable academic context and traceable inspiration pathways. These shortcomings limit the credibility and practical value of LLM-generated ideas within scientific contexts that prioritize trustworthiness, traceability, and reproducibility.

To mitigate these issues, Retrieval-Augmented Generation (RAG)\cite{gao_synergizing_2025,chen_dense_2024} and Microsoft's GraphRAG\cite{edge_local_2025,han_retrieval-augmented_2025} provide LLMs with contextual evidence retrieved from external high-quality knowledge bases, which significantly reduces hallucinations and enhances factual substantiation. Related graph-augmented approaches to query-focused summarisation further underline the value of structured evidence for controllable generation. However, simulating the \emph{origins of inspiration} and the \emph{pathways of insight} remains inadequate.

To address these challenges, this paper proposes \textbf{GYWI}, a hybrid LLM framework for generating research ideas. The system aims to simulate the process of scientific inspiration, with the task objective of automatically generating novel research ideas based on a given target paper. It comprises the following three key components:

     \textit{1) Author-centered Knowledge Graph and Inspiration Source Sampling}: Constructs an author knowledge graph based on co-authorship relationships, thematic associations, and community structures (author--paper perspective), drawing on academic graph resources; and designs sampling algorithms utilising neighbouring hops (\emph{exploitation}) and random hops (\emph{exploration}) to form an inspiration-source collection.

    \textit{2) Hybrid Retrieval via RAG and GraphRAG}: Employs a ``depth and breadth'' hybrid retrieval mechanism: RAG supplies fine-grained local evidence, while GraphRAG contributes graph-level global connections (entities as nodes, relations as edges), generating the contextual framework.
    
    \textit{3) Prompt optimization Strategy} Based on Reinforcement Learning Principles: Automatically fine-tunes generation strategies within the hybrid context to optimise use of retrieved evidence, normalise output structure, and enhance the relevance and reliability of generated ideas (in line with recent progress in reasoning-oriented prompting and multi-step idea development).

For evaluation, we constructed a benchmark dataset based on arXiv papers from 2018--2023 to support multi-dimensional assessment (automated evaluation of The Idea Multiple-Choice Evaluation (IMCQ) objective questions, LLM-based scoring, human evaluation, and semantic-space visualization). 
The main contributions of this paper are:

\begin{itemize}
    % \item author-centered knowledge modelling for traceable ideation: We propose an author-centered knowledge graph combined with an inspiration-source sampling algorithm, integrated with retrieval-augmented generation techniques to generate research ideas, thereby enhancing scientific credibility and traceability.
     \item author-centered knowledge modeling for controllable and traceable ideation: We propose an external author-centered knowledge graph combined with an inspiration-source sampling algorithm, integrated with retrieval-augmented generation techniques to generate research ideas under a controllable context with traceable inspiration paths, thereby enhancing scientific credibility and traceability.
    \item Prompt optimization with reinforcement learning principles: We introduce a prompt-optimization strategy that guides LLMs to perform standardised processing---such as structural adjustment and information filtering---over retrieved content, improving the effectiveness of retrieval-augmented generation for scientific idea generation tasks.
    \item Evaluation dataset and framework: We construct an arXiv-based assessment dataset (2018--2023) and a comprehensive evaluation methodology encompassing subjective human assessment, automated LLM scoring, and objective multiple-choice (IMCQ) evaluation.
\end{itemize}

\section{Related Work}

With the rapid development of artificial intelligence, the use of artificial intelligence to assist scientific research tasks has become a popular research direction. In the early years, researchers began to use knowledge graphs to construct large-scale academic association networks, such as ArnetMiner~\cite{tang_arnetminer_2008} and the subsequent MAG~\cite{microsoft_academic_graph} and OAG~\cite{oag} proposed by Microsoft, providing researchers with comprehensive information retrieval tools. However, these efforts mainly focused on indexing and retrieval, and did not directly involve the process of generating scientific research ideas.

With the development of LLMs in recent years, researchers have begun to explore the possibility of using LLMs as AI scientists. For example, Lu et al.~\cite{lu_ai_2024} discussed the possibility of using LLMs as AI scientists. Similarly, the works of Baek et al.~\cite{baek_researchagent_2025}, Li et al.~\cite{li_chain_2024} and Radensky et al.~\cite{radensky_scideator_2025} also used LLMs to generate scientific research ideas through multi-round dialogue and iterative feedback. However, these methods generally have two limitations. First, they lack academic factual support, making them prone to ‘hallucinations' and resulting in content with insufficient credibility. Second, they cannot incorporate deep academic connections during the generation process, making it difficult to simulate the ‘flash of inspiration' process that researchers experience in real-world scenarios, and the generated content often lacks novelty and academic relevance. Addressing these issues, it is crucial to systematically evaluate and fundamentally improve the quality of generated ideas. From early manual evaluations~\cite{sun_scieval_2024} to strategies combining LLM-assisted assessments, representative work such as AI Idea Bench~\cite{qiu_ai_2025} has designed multi-dimensional subjective and objective evaluation methods to systematically measure the quality of generated ideas. Despite the increasing maturity of evaluation methods, the fundamental issues on the generation side remain unresolved.

% Existing research indicates that current studies have failed to systematically integrate the structured knowledge of academic networks with the generative capabilities of large language models (LLMs), resulting in generated research ideas that lack deep academic connections. The ideas produced exhibit significant shortcomings in terms of innovation, relevance, and factual basis. In response to these challenges, this paper proposes the GYWI system, which integrates academic relationship networks, retrieval-enhanced generation, and prompt optimization strategies. The system aims to enhance the academic relevance, innovation, and feasibility of generated ideas and validates its effectiveness through a multi-dimensional evaluation method.
LLM-based idea generation often lacks controllable academic context and traceable inspiration pathways. Existing research also indicates that current studies have not systematically integrated the structured knowledge of academic networks with the generative capabilities of large language models (LLMs), resulting in generated research ideas that lack deep academic connections and show shortcomings in innovation, relevance, and factual grounding. To bridge this gap, this paper proposes the GYWI system, which builds an external author-centered knowledge graph with inspiration-source sampling, forms a hybrid context via a depth and breadth retrieval mechanism (RAG + GraphRAG), and applies a prompt optimization strategy to guide the LLM. The system aims to enhance the academic relevance, innovation, and feasibility of generated ideas, and validates its effectiveness through a multi-dimensional evaluation method.

\section{The proposed Approach: GYWI for Idea Generation}
In this section, we present the overall framework and implementation of GYWI, which comprises four modules: an author knowledge graph, a retrieval-enhanced generation module, a prompt-optimization module, and an LLM generator.
\begin{figure*}[htbp]
\centering
\includegraphics[width=\textwidth]{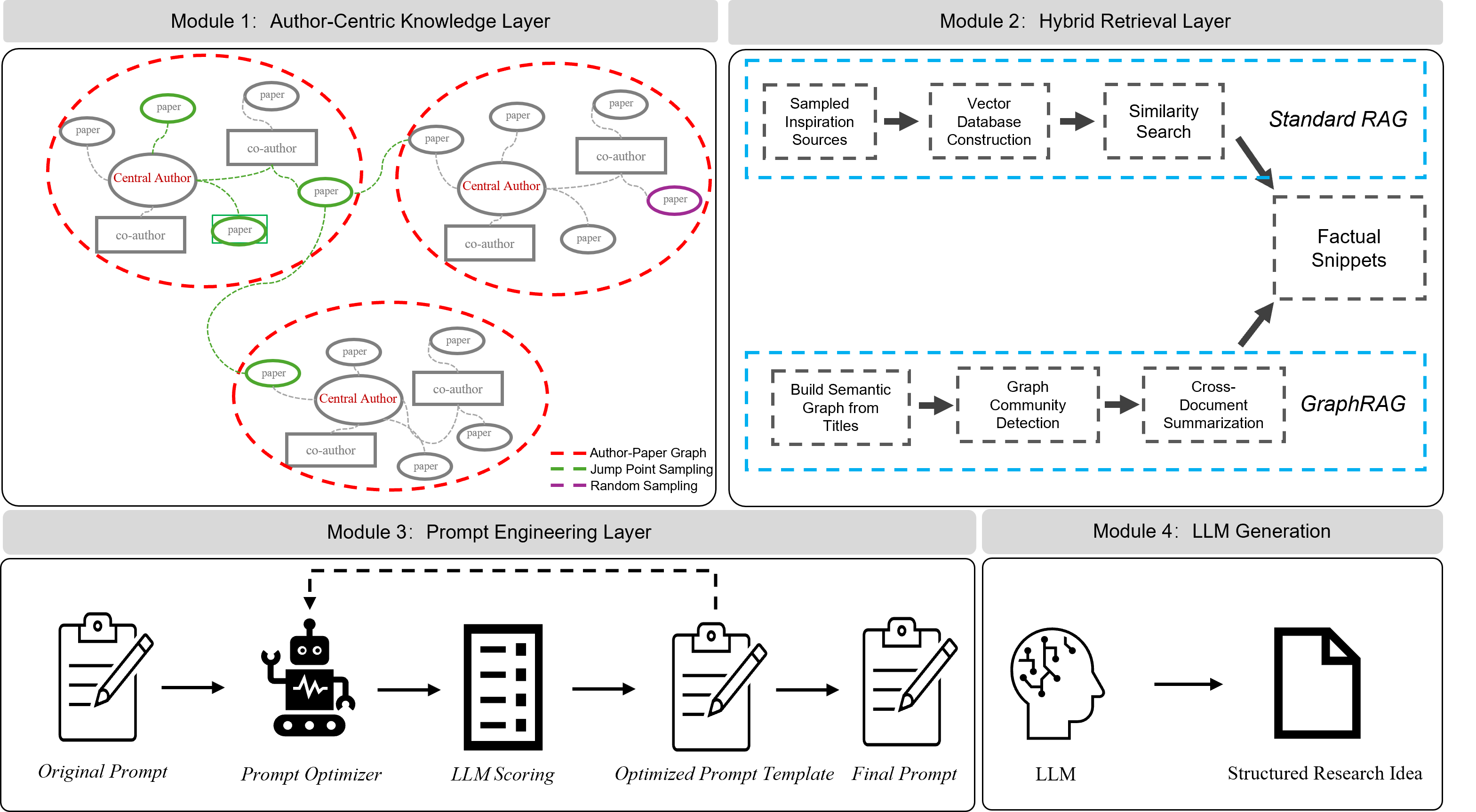}
\caption{\textbf{Author-guided scientific research idea generation system}\\
 \small Note: This system combines author knowledge graphs with hybrid retrieval methods to generate scientific research ideas.}
 \label{fig:Framework}
\end{figure*}
% We build GYWI around an author-centered knowledge graph and use it with a sampling algorithm and retrieval-augmented techniques to obtain verifiable factual evidence from real research papers. 

We first build an author--paper bipartite graph $G=(A^{*}\cup P^{*},E)$ from full papers and metadata.Concretely, we construct an author–paper bipartite graph from full papers and metadata, detect communities, and collect candidate inspiration sources via adjacent (exploitation) and random (exploration) hops. On top of these sources, a hybrid retrieval scheme (RAG for depth, GraphRAG for breadth) extracts evidence and assembles a controllable context with traceable provenance. Finally, the prompt optimization module guides the base LLM to organize the retrieved context and generate research ideas (see Fig.~\ref{fig:Framework}).

\begin{table}[t]
\centering
\caption{Semantic similarity between co-authors vs. non-collaborative pairs}
\label{tab:semantic_similarity}
\begin{tabular}{lcc}
\toprule
Author Pair Type & Avg. Semantic Similarity \\
\midrule
Co-authors & 0.0604 \\
Non-collaborative Pairs & 0.0134 \\
\bottomrule
\end{tabular}
\end{table}

\subsection{author-centered Knowledge Graph}
To simulate the propagation path of academic inspiration, we constructed an academic knowledge graph based on central authors and designed a set of controllable inspiration sampling algorithms on this basis.

\subsubsection{Knowledge Graph Construction}
\begin{figure}[htbp]
  \centering
  \includegraphics[width=0.85\linewidth]{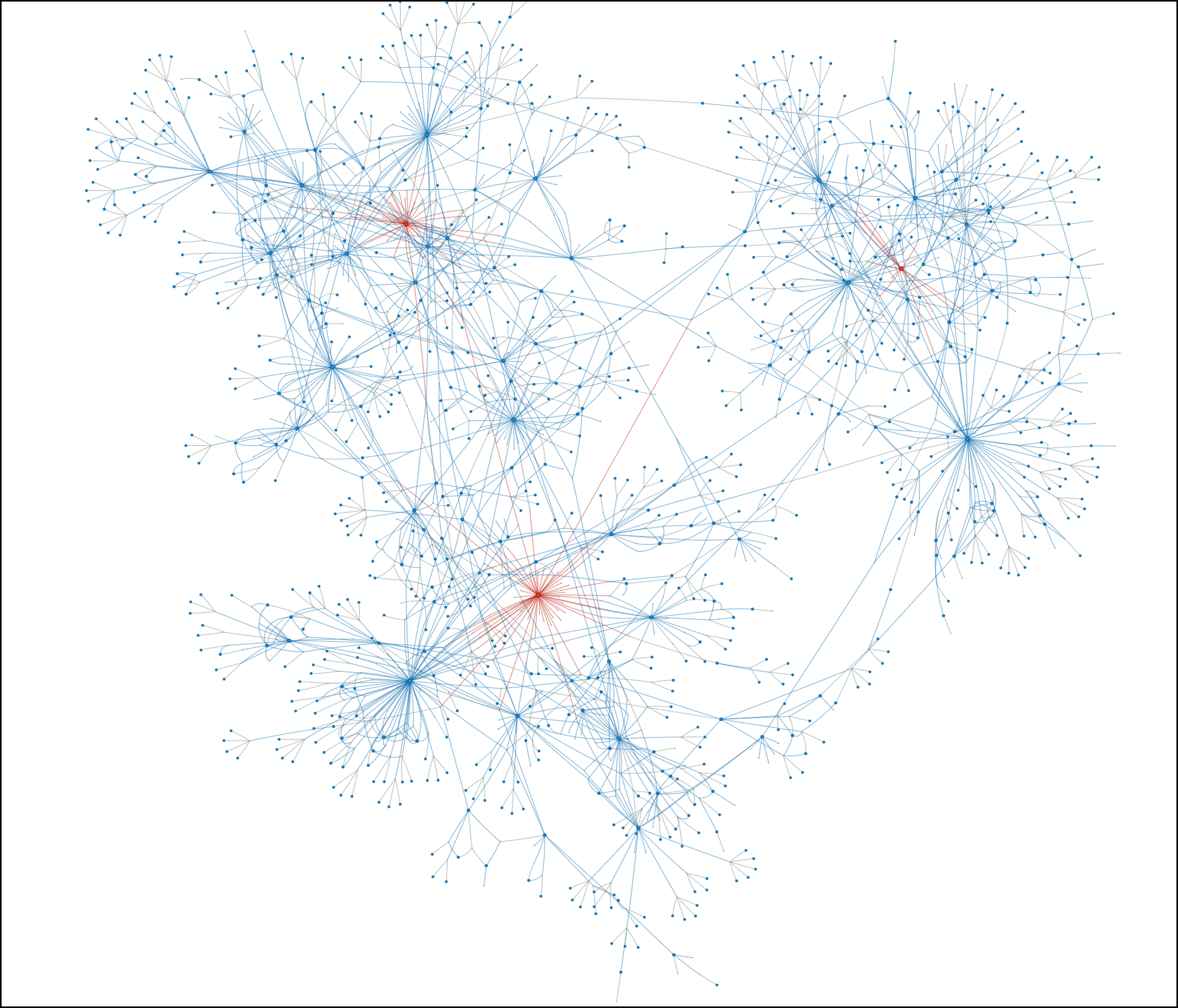}
  \caption{\textbf{Author Knowledge Graph (generated example).}
  This auto-generated graph illustrates the collaboration network derived from co-authorships around a target paper.}
  \label{fig:author_knowledge_graph}
\end{figure}

To simulate the propagation paths of academic inspiration, we constructed an author-centered knowledge graph. In reality, most highly productive authors typically collaborate with a stable group of researchers on similar topics, publishing multiple papers together and forming research communities. We analysed paper data from the artificial intelligence subfield of arXiv (2018-2023) and found that collaborators have strong associations in research directions by comparing the semantic similarity of paper titles between collaborators and non-collaborators (Table~\ref{tab:semantic_similarity}). Based on this finding, we constructed an author knowledge graph. First, we selected a set of paper authors as the seed set $A_{\text{seed}}$ (for example, in our experiment, we selected scholars such as Yann LeCun, Geoffrey Hinton, and Kaiming He as examples); then the algorithm mined the collaborative academic networks of these central authors and used the Louvain community algorithm\cite{blondel2008fast} to identify highly correlated core academic communities; Finally, the algorithm constructs an author-paper bipartite graph $G$ based on this community and the author set $A^*$ and paper node set $P^*$, where each edge represents the relationship between an author and a paper (Fig.~\ref{fig:author_knowledge_graph}).

\begin{example}[Author and Paper Sets]\label{ex:AstarPstar}
If $A_{\text{seed}}=\{\text{Author}_1,\text{Author}_2,\text{Author}_3\}$, then after community detection and merging, $A^{*}$ contains their frequent co-authors within the selected communities; $P^{*}$ collects all papers authored by $A^{*}$. This expands the inspiration scope from seed authors to their community-level collaborators.
\end{example}

\begin{algorithm}[h]
\caption{Author Knowledge Graph Construction}
\label{alg:graph_construction}
\begin{algorithmic}[1]
\Require Seed author set $A_{\text{seed}}$, paper corpus $P_{\text{corpus}}$, community detection algorithm $F_{\text{community}}$
\Ensure Author-paper bipartite graph $G=(A^* \cup P^*, E)$
\State Extract paper set $P_{\text{seed}} \gets \{p \in P_{\text{corpus}} \mid \text{author}(p) \cap A_{\text{seed}} \neq \emptyset \}$.
\State Construct initial co-author graph $G_{\text{co-author}}$ based on co-authorships in $P_{\text{seed}}$.
\State Apply $F_{\text{community}}$ to $G_{\text{co-author}}$ to obtain communities $C = \{C_1, C_2, \dots\}$.
\State Identify communities containing seed authors: $C' \gets \{C_i \mid C_i \cap A_{\text{seed}} \neq \emptyset\}$.
\State Merge authors in these communities to form the core author set: $A^* \gets \bigcup_{C_i \in C'} C_i$.
\State Extract all papers $P^* \gets \{p \in P_{\text{corpus}} \mid \text{author}(p) \cap A^* \neq \emptyset \}$.
\State Construct bipartite graph $G=(A^* \cup P^*, E)$ where $E$ represents authorship links.
\State \Return $G$
\end{algorithmic}
\end{algorithm}

\subsubsection{Inspiration Source Sampling Algorithm}
In reality, inspiration often involves both exploitation and exploration, meaning that academic inspiration may come from highly related academic articles or ideas, or it may come from any knowledge point—even if that knowledge point appears to be unrelated to the target. To simulate this phenomenon, we designed an \textit{inspiration association} and \textit{inspiration jump} sampling algorithm to provide knowledge sources for subsequent inspiration generation. The algorithm primarily consists of two parts:
\begin{itemize}
    \item \textbf{Adjacent Jump Point Sampling:} This part focuses on in-depth exploration of related knowledge points. The algorithm identifies N-order author groups adjacent to $p_{\text{target}}$ in the knowledge graph $G$, then samples from these authors' paper collections to obtain content closely related to the target topic.
    \item \textbf{Random Jump Point Sampling:} This component simulates the ‘eureka moment,' which is a moment of inspiration. The algorithm randomly samples from the entire paper collection in the knowledge graph, introducing diverse or even seemingly unrelated topics to provide possibilities for stimulating LLM associations.
\end{itemize}
\begin{example}[Adjacent Sampling Around Target Paper]\label{ex:ptarget-adjacent}
Let $p_{\text{target}}$ be a recent arXiv paper in the target domain. 
For $N{=}1$, the adjacent author set $A_{\text{adj}}$ contains direct co-authors of the target paper's authors; 
$P_{\text{adj}}$ are their publications excluding $p_{\text{target}}$. 
Sampling $k_{\text{adj}}$ from $P_{\text{adj}}$ yields highly related evidence snippets.
\end{example}

\begin{algorithm}[h]
\caption{Inspiration Source Sampling Algorithm}
\label{alg:jump_point_sampling}
\begin{algorithmic}[1]
\Require 
    Knowledge graph $G = (A^* \cup P^*, E)$;\\
    Target paper $p_{\text{target}}$;\\
    Neighborhood order $N$;\\
    Number of adjacent samples $k_{\text{adj}}$;\\
    Number of random samples $k_{\text{rand}}$
\Ensure 
    Set of inspiration source papers $P_{\text{jump}}$
\vspace{0.5em}
\State $P_{\text{jump}} \gets \emptyset$
\Comment{\textit{Adjacent Jump Point Sampling (Exploitation)}}
\State Identify $A_{\text{adj}} \gets$ the set of authors within $N$-hop neighborhood of $p_{\text{target}}$ in $G$
\State Retrieve $P_{\text{adj}} \gets$ all papers authored by $A_{\text{adj}}$
\State Sample $k_{\text{adj}}$ papers from $P_{\text{adj}} \setminus \{p_{\text{target}}\}$ and add to $P_{\text{jump}}$
\vspace{0.5em}
\Comment{\textit{Random Jump Point Sampling (Exploration)}}
\State Define $P_{\text{candidates}} \gets P^* \setminus (P_{\text{adj}} \cup \{p_{\text{target}}\})$
\State Randomly sample $k_{\text{rand}}$ papers from $P_{\text{candidates}}$ and add to $P_{\text{jump}}$
\vspace{0.5em}
\State \Return $P_{\text{jump}}$
\end{algorithmic}
\end{algorithm}

\paragraph{Parameterisation}
$N$ (hop order): neighborhood depth on $G$;
$k_{\text{adj}}$: number of adjacent samples (\textit{exploitation});
$k_{\text{rand}}$: number of random samples (\textit{exploration}).
These hyperparameters control the trade-off between factual grounding (high relevance) and exploratory novelty (high diversity).

The number of samples for associated jump points $k_{\text{adj}}$ and the number of samples for random jump points $k_{\text{rand}}$ are both adjustable hyperparameters. By adjusting the ratio of the two hyperparameters, we can precisely control the corpus input to the LLM, thereby balancing the factual basis (high relevance) and innovative exploration (high diversity) of the final generated scientific ideas.

\subsection{Hybrid Retrieval for Depth and Breadth}
Based on the inspiration nodes extracted above, we propose a hybrid retrieval strategy that combines RAG and GraphRAG to extract the academic knowledge we need.
\begin{itemize}
    \item \textbf{Depth Content Retrieval via RAG:} Based on the sampling algorithm mentioned above, we perform precise semantic retrieval on relevant nodes in the graph, providing the LLM with the necessary, fine-grained ‘factual fragments' and ‘innovative fragments' required for idea generation, thereby enhancing the depth of the input content to the LLM.
    \item \textbf{Breadth Content Retrieval via GraphRAG:} In addition to RAG, we employ GraphRAG technology to further enhance the structure and diversity of generated content. We no longer rely on sampling algorithms but directly use GraphRAG to capture the macro-level structure of the academic network and core relationships between papers through paper titles, generating academic reports based on target papers. This effectively addresses the limitation of RAG, which struggles to retrieve structural and entity relationships between documents in generated content, thereby enhancing the breadth of input content for the LLM.
\end{itemize}

\begin{example}[Minimal illustrations of the hybrid retrieval]\label{ex:hybrid-min}
(a) RAG depth retrieval
\begin{exbox}
\footnotesize\ttfamily
[Chunk 1] From Paper 2010.13337 (self, Score: 1.0000) ...\par
[Chunk 9] From Paper 2110.07858 (neighbor, Score: 0.6218) ...\par
[Chunk 10] From Paper 1911.09665 (random, Score: 0.6600) ...
\end{exbox}

(b) GraphRAG breadth summarisation
\begin{exbox}\footnotesize\ttfamily
Title: ...\\
\#\# Overview of Key Methods\\
- ...\\
\#\# Implementation Insights\\
- ...\\
\#\# Performance Considerations\\
- ...
\end{exbox}
\end{example}

Through this \textit{depth mining + breadth coverage} strategy, the hybrid retrieval module \emph{can} improve the relevance, reliability, and academic feasibility of the generated research ideas.

\subsection{Adaptive Prompt Optimization for Idea Generation}
To enable LLM to effectively utilize the complex information retrieved through hybrid retrieval, we designed a prompt optimization strategy based on reinforcement learning principles. We found that simple prompt templates used in the past tended to overload the system with information after retrieving complex information through hybrid retrieval, making it difficult to focus. We use a \texttt{Prompt\_Optimizer} constructed by the LLM to automatically optimise prompts, enabling the LLM to better adapt to the complexity of hybrid retrieval information.

\subsubsection{Prompt Optimization Strategy}
We optimize the prompt template $P$ using LLM-generated scores in five dimensions:
\[
\mathbf{S}(P)=\bigl[
s_{\text{novelty}},\;
s_{\text{feasibility}},\;
s_{\text{clarity}},\;
s_{\text{relevance}},\;
s_{\text{significance}}
\bigr],
\]
and maximize the average score
\[
P^{*}=\arg\max_{P}\;\bar{S}(P),\qquad 
\bar{S}(P)=\frac{1}{5}\sum_{i=1}^{5}s_{i}.
\]

% To ensure robust evaluation, we randomly sample 10 topics per iteration, reuse each prompt $R=5$ times, and estimate performance via the sample mean $\bar{s}$ and standard error:
% \[
% SE = \frac{s}{\sqrt{N_{\text{eval}}}}, \quad \text{CI}_{95\%} = \bar{s} \pm t^{0.975}_{N_{\text{eval}} - 1} \cdot SE.
% \]
% A prompt is updated only if its improvement is statistically significant.
To ensure robust evaluation, in each iteration we sample $N_{\text{topic}}$ topics and reuse each prompt $R$ times; the total number of evaluations is
\[
N_{\text{eval}} = N_{\text{topic}} \times R.
\]
We estimate performance using the sample mean $\bar{s}$ and standard error
\[
\mathrm{SE} = \frac{s}{\sqrt{N_{\text{eval}}}},
\qquad
\mathrm{CI}_{1-\alpha} = \bar{s} \pm t^{\,1-\alpha/2}_{\,N_{\text{eval}}-1}\cdot \mathrm{SE},
\]
where $s$ is the sample standard deviation and $t^{\,1-\alpha/2}_{\,\nu}$ is the critical value of the Student's $t$-distribution with $\nu$ degrees of freedom. A prompt is updated only if the observed improvement over the current best is statistically significant at level $\alpha$.

In our experiments we set $N_{\text{topic}}=10$, $R=5$, and $\alpha=0.05$. These values are evaluation settings for statistical precision and do not constitute model hyperparameters.

\noindent Notation.
\begin{center}
\footnotesize
\begin{tabular}{ll}
\hline
Symbol & Meaning \\
\hline
$N_{\text{topic}}$ & topics/iter (10) \\
$R$ & reuses/topic (5) \\
$N_{\text{eval}}$ & total evals, $N_{\text{topic}}R$ \\
$\bar{s}$ & sample mean score \\
$s$ & sample std.\ dev. \\
$\mathrm{SE}$ & $s/\sqrt{N_{\text{eval}}}$ \\
$\mathrm{CI}_{1-\alpha}$ & $\bar{s}\pm t^{1-\alpha/2}_{\,N_{\text{eval}}-1}\mathrm{SE}$ \\
$t^{1-\alpha/2}_{\nu}$ & $t$ critical value (df $\nu$) \\
$\alpha$ & significance level (0.05) \\
\hline
\end{tabular}
\end{center}

The \texttt{Prompt\_Optimizer} uses previous prompts, their scores, and structural constraints (e.g., XML format) to refine both the task description and retrieved content. This iterative update process is analogous to policy optimization in reinforcement learning, where the prompt acts as the policy, modifications as actions, and scores as rewards.

% Ultimately, through multiple rounds of ‘\textit{sampling-scoring-feedback}' iterations, our prompt template gradually learned how to better organize and guide the retrieved information, thereby consistently generating high-quality scientific ideas. The original and optimised prompts are detailed \textit{in} Appendix E.

Ultimately, we run iterative prompt optimization for 5--10 rounds and stop when the average LLM score no longer improves for two consecutive rounds. Compared with the original prompt (single-document, fixed schema, no explicit ties to retrieved evidence), the optimised prompt consumes a two-layer context (GraphRAG summary plus selected RAG chunks), requires synergy or gap-filling, and adds explicit quality criteria and provenance-aware fields, enabling the LLM to organize and use evidence more effectively. The original, intermediate, and optimised prompts are provided in Appendix~\ref{appendix:prompts}.

\section{Experiments}
In this section, we present the experimental design and result analysis. 
Subsection~\ref{subsec:eval} introduces our comprehensive evaluation methodology, 
and Subsection~\ref{subsec:results} reports the final results and analyses.

\subsection{Evaluation Method}\label{subsec:eval}

To assess the effectiveness of GYWI in generating research ideas, we designed four types of experiments: matching evaluation, LLM-based scoring, human scoring, and semantic space visualization analysis. Additionally, we selected 99 target papers randomly sampled from our 1{,}000-paper evaluation corpus, and asked four models---GPT\textendash 4o, DeepSeek\textendash V3, Qwen3\textendash 8B, and Gemini~2.5~Flash---to generate ideas for them, comparing the results with our system.

\subsubsection{Dataset construction}

\begin{figure*}[htbp]
    \centering
    \begin{subfigure}[t]{0.40\textwidth}
        \centering
        \includegraphics[width=\linewidth]{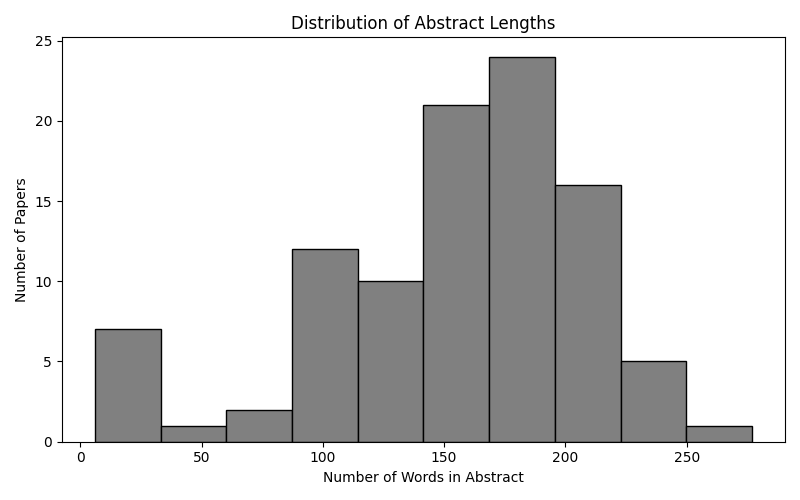}
        \caption{Distribution of abstract lengths in the dataset.}
        \label{fig:abstract_len}
    \end{subfigure}
    \hfill
    \begin{subfigure}[t]{0.40\textwidth}
        \centering
        \includegraphics[width=\linewidth]{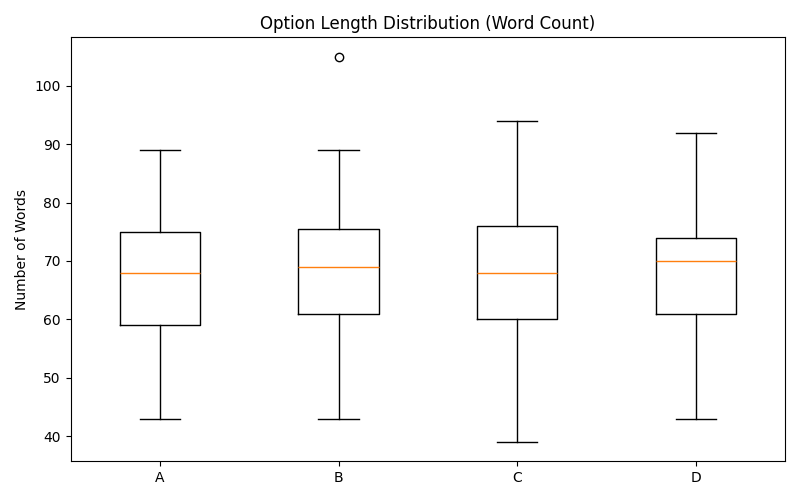}
        \caption{Boxplot of word counts in IMCQ options (A–D).}
        \label{fig:option_len}
    \end{subfigure}
    \caption{\textbf{Dataset Statistics.} (a) Abstracts are mostly between 120–180 words, providing sufficient context for motivation extraction. (b) IMCQ options are length-balanced, avoiding answer bias.}
    \label{fig:dataset_stats}
\end{figure*}

To assess the model's actual performance in generating scientific research ideas, we referenced the method used in AI Idea Bench 2025~\cite{qiu_ai_2025} to construct an evaluation dataset. We randomly selected 1,000 papers from the artificial intelligence subfield of arXiv (2018-2023). For each paper, we first used DeepSeek-V3 to extract its true motivation, which served as the \textbf{correct answer}. Then, we utilized Sentence-BERT to select three distractors from the AI Idea Bench dataset that were semantically similar to the correct answer but had different themes. Finally, each multiple-choice question (IMCQ) was composed of one correct answer and three distractors, forming a standard multiple-choice question. This dataset provides the data foundation for our experiments in Sections~\ref{subsec:imcq} and~\ref{subsec:comparison}, with its statistical information summary shown in Figure~\ref{fig:dataset_stats}. The dataset has been made publicly available at \url{https://github.com/Pengzhen97/GYWI} in the file \texttt{updated\_mcq\_with\_id\_abstract.json}.

\subsubsection{Multi-dimensional Scoring Assessment}

To assess the overall quality of research ideas generated, we employ a comprehensive scoring system based on five dimensions: novelty, feasibility, clarity, relevance, and significance. Each dimension is scored on a scale from 1 (lowest) to 10 (highest).

\textbf{LLM-based scoring:} We utilise DeepSeek-V3 for LLM automated scoring, conducting blind evaluations of research ideas from five systems using standardised scoring prompts.

\textbf{Human scoring:} We used the same method to invite five graduate students from our university, who have relevant domain backgrounds, participated in a blind evaluation. They possess relevant domain knowledge and are capable of assessing the generated research ideas. To ensure a fair and unbiased assessment, the evaluation was conducted in a blind manner, where the identity of the source model for each generated idea was withheld from the evaluators. Details are provided in Appendix~\ref{appendix:materials}.

For both evaluation methods, the final score is the average of the scores across all dimensions, effectively assessing the system's overall quality and performance in generating research ideas.

\subsection{Experimental Results and Analysis}\label{subsec:results}
In this section, we conduct experimental evaluation and analysis following the methodology in Subsection~\ref{subsec:eval}, and perform an ablation study of GYWI's components using IMCQ multiple-choice questions to assess its capability in generating scientific research ideas. The evaluation prompt is provided in Appendix~\ref{appendix:evaluation_prompt}. Next, we will perform semantic visualization of the generated content.

\subsubsection{Idea Multiple-Choice Question (IMCQ)}
\label{subsec:imcq}
% \begin{figure}[htbp]
%     \centering
%     \begin{subfigure}[t]{0.49\textwidth}
%         \centering
%         \includegraphics[width=\linewidth]{Accuracy.png}
%         \caption{ Accuracy of GYWI based on different LLMs in the IMCQ evaluation (4.2.1).}
%         \label{fig:imcq_bar}
%     \end{subfigure}
%     \hfill
%     \begin{subfigure}[t]{0.49\textwidth}
%         \centering
%         \includegraphics[width=\linewidth]{bar_chart_compare.png}
%         \caption{IMCQ Accuracy Improvement by Module (4.2.2).}
%         \label{fig:imcq_gain}
%     \end{subfigure}
%     \caption{\textbf{IMCQ Evaluation Results: Accuracy by LLM and Improvement by Added Modules.}}
%     \label{fig:imcq_test}
% \end{figure}

\begin{figure}[htbp]
    \centering
    \includegraphics[width=0.8\linewidth]{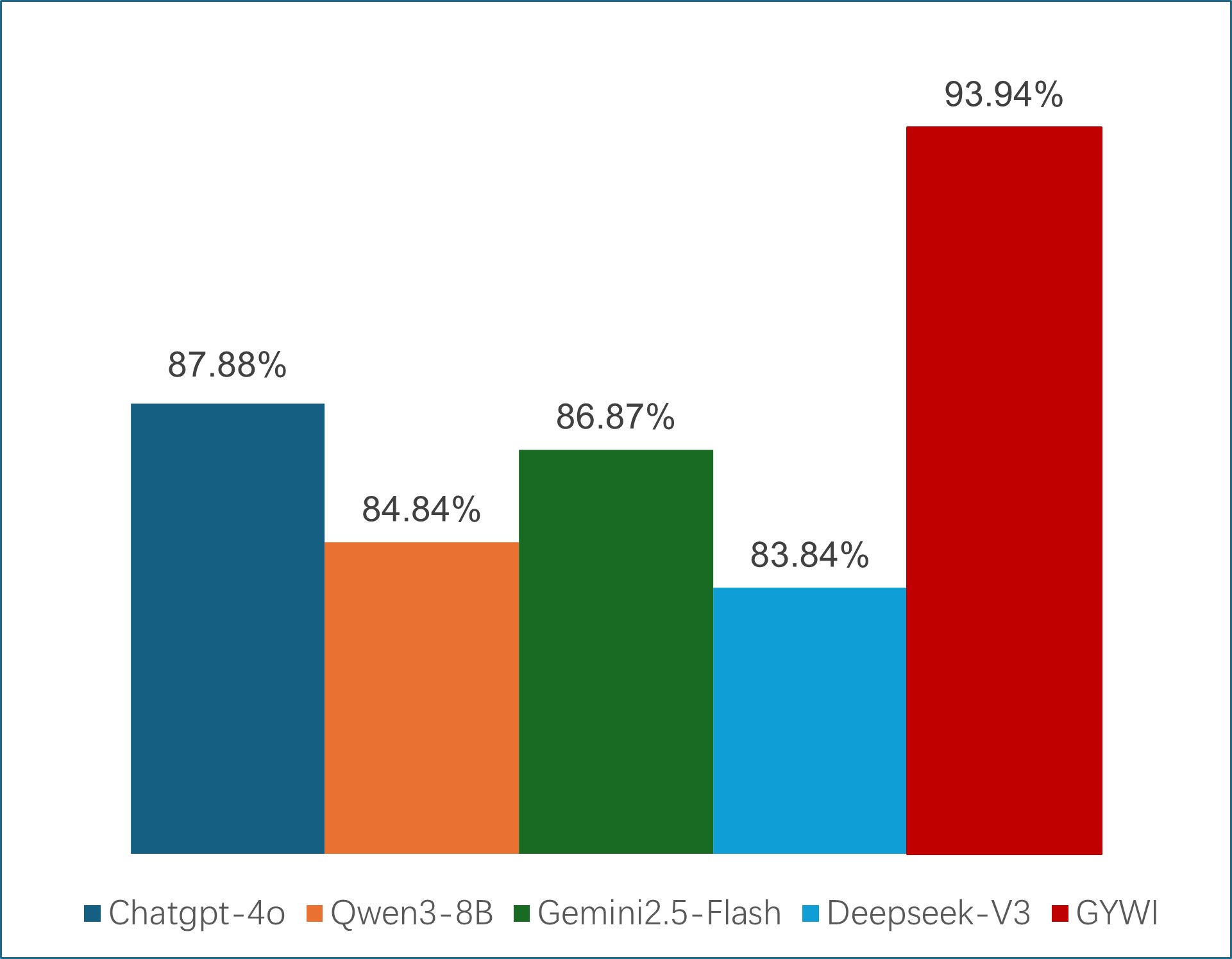}
    \caption{Accuracy of GYWI based on different LLMs in the IMCQ evaluation.}
    \label{fig:imcq_bar}
\end{figure}

\begin{figure}[htbp]
    \centering
    \includegraphics[width=0.8\linewidth]{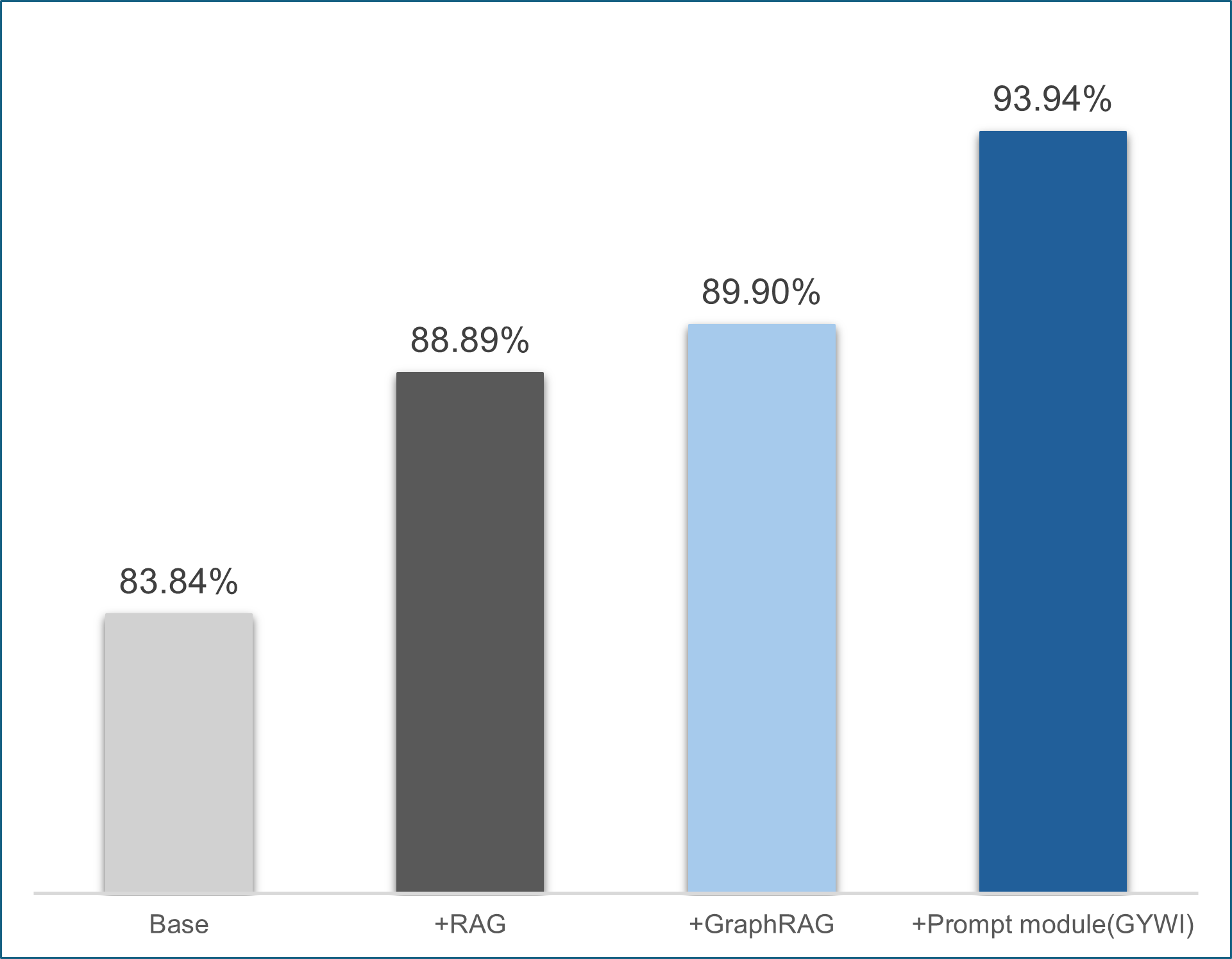}
    \caption{IMCQ Accuracy Improvement by Module.}
    \label{fig:imcq_gain}
\end{figure}

The experimental results are shown in Fig.~\ref{fig:imcq_bar}. Using DeepSeek\textendash V3 as the base LLM, our model (GYWI) attains the highest accuracy of 93.94\% on this task, an absolute improvement of more than 10 percentage points over the DeepSeek\textendash V3 baseline. It also significantly outperformed all other LLMs, indicating that retrieval-based generation enhancement under the author knowledge graph framework can effectively align the generated research motivations with the original research topic, thereby significantly improving the alignment between the generated content and the actual motivations. For an example of an IMCQ question, please refer to Appendix~\ref{appendix:imcq_example}.

\subsubsection{Ablation Study}
\label{subsec:comparison}

To evaluate the necessity and effectiveness of each module in this system for the final outcome, we designed a series of experiments using IMCQ multiple-choice questions to assess accuracy and compare the three core components of the system. Considering that the author knowledge graph is paired with retrieval-based generation, we conduct an ablation on a single backbone LLM, \textbf{DeepSeek-V3}: starting from the base model (no retrieval), we incrementally enable RAG, GraphRAG, and the prompt-optimization module, and report the change in accuracy.

As shown in the results (Fig.~\ref{fig:imcq_gain}), each module has improved the final accuracy to varying degrees. Specifically, the accuracy improved by 5.05 percentage points after adding RAG, by 1.01 percentage points after adding GraphRAG, and by 4.04 percentage points after adding the prompt optimization module. Essentially, we believe that the greatest improvement to the model comes from the combination of the author knowledge graph and the retrieval-generation module, while the modification of the prompt template is merely to help the model better understand the retrieved content.

\subsubsection{Automated Scoring by LLM}
% \begin{figure}[htbp]
%     \centering
%     \resizebox{0.75\textwidth}{!}{
%     \begin{minipage}{0.45\linewidth}
%         \centering
%         \includegraphics[width=\linewidth]{rader.png}
%     \end{minipage}%
%     \hfill
%     \begin{minipage}{0.53\linewidth}
%         \centering
%         \includegraphics[width=\linewidth]{LLM_eva.png}
%     \end{minipage}
%     } % 结束resizebox
%     \caption{\textbf{Automated Scoring by LLM}\\
%     \small Radar chart and table summarising LLM evaluation scores for clarity, feasibility, novelty, relevance, and significance.}
%     \label{fig:subjective}
% \end{figure}

\begin{figure}[htbp]
    \centering
    \includegraphics[width=0.8\linewidth]{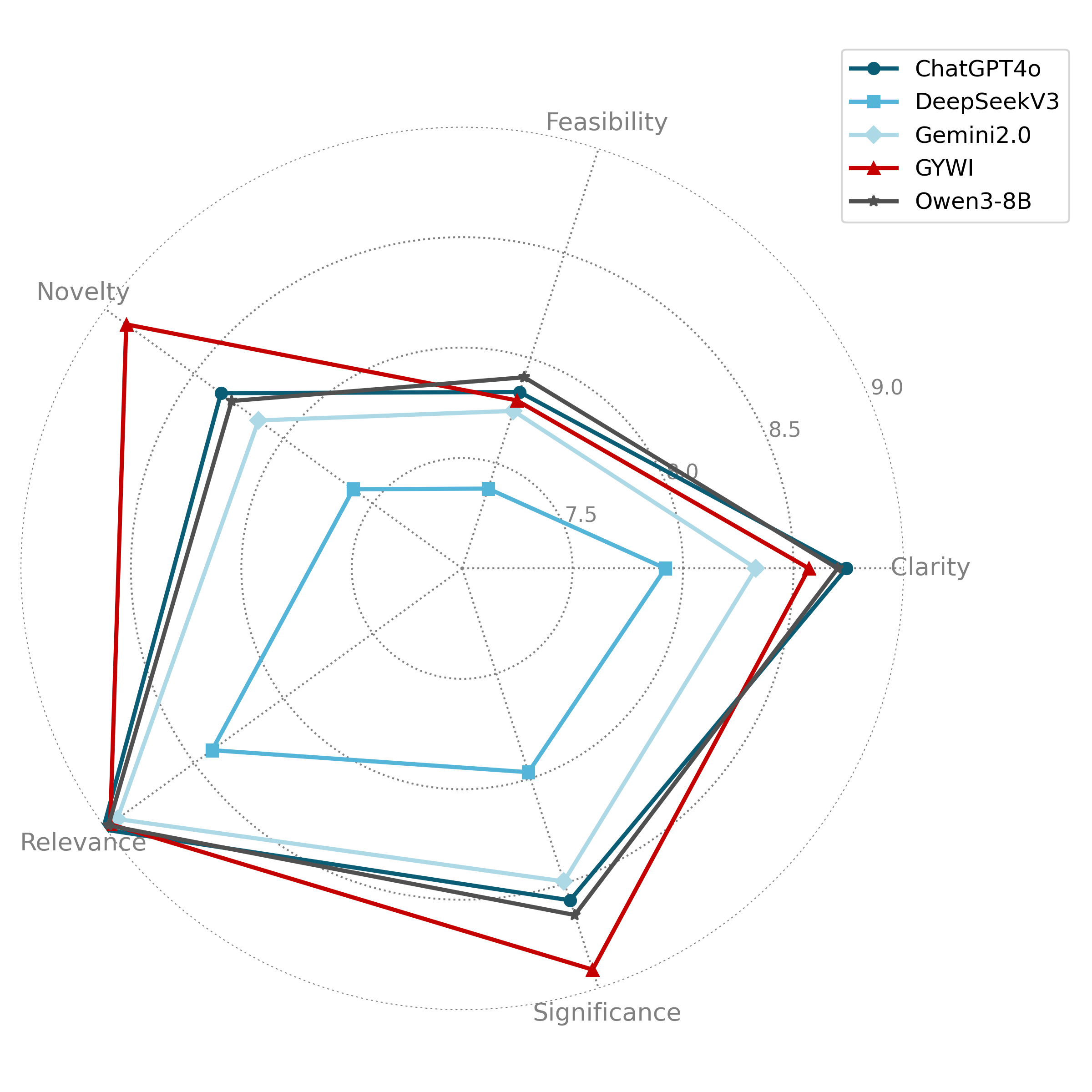}
    \caption{LLM evaluation of generated ideas.}
    \label{fig:llm_radar}
\end{figure}

The experimental results are shown in Fig.~\ref{fig:llm_radar}. The GYWI model performed best overall, achieving the highest average score (8.626) among all models and the highest scores in the two key scientific creativity indicators of Novelty and Significance. When compared to the baseline model DeepSeek-V3, significant improvements were observed across all dimensions, indicating that the knowledge graph-enhanced retrieval and prompt optimization mechanisms effectively enhance the quality, accuracy, and academic potential of the generated content, with the model demonstrating a significant advantage in terms of relevance.

% \begin{figure}[htbp]
%     \centering
%     \begin{subfigure}[t]{0.49\textwidth}
%         \centering
%         \includegraphics[width=\linewidth]{雷达图.png}
%         \caption{LLM evaluation of generated ideas }
%         \label{fig:llm_radar}
%     \end{subfigure}
%     \hfill
%     \begin{subfigure}[t]{0.49\textwidth}
%         \centering
%         \includegraphics[width=\linewidth]{human_雷达图.png}
%         \caption{Human evaluation of generated ideas}
%         \label{fig:human_radar}
%     \end{subfigure}
%     \caption{\textbf{Comparative Evaluation of Idea Quality by LLM and Humans.}}
%     \label{fig:evaluation_radar_charts}
% \end{figure}

\subsubsection{Human Evaluation}
% \begin{figure}[htbp]
%     \centering
%     \resizebox{0.75\textwidth}{!}{
%     \begin{minipage}{0.45\linewidth} 
%         \centering
%         \includegraphics[width=\linewidth]{human_雷达图.png}
%     \end{minipage}%
%     \hfill
%     \begin{minipage}{0.53\linewidth}  
%         \centering
%         \includegraphics[width=\linewidth]{eva_human.png}
%     \end{minipage}
%     }
%     \caption{\textbf{Human Evaluation}\\
%     \small Human evaluation across the same five dimensions as the automated scoring shows GYWI ranking highest in all.}
%     \label{fig:human_evaluation}
% \end{figure}

\begin{figure}[htbp]
    \centering
    \includegraphics[width=0.8\linewidth]{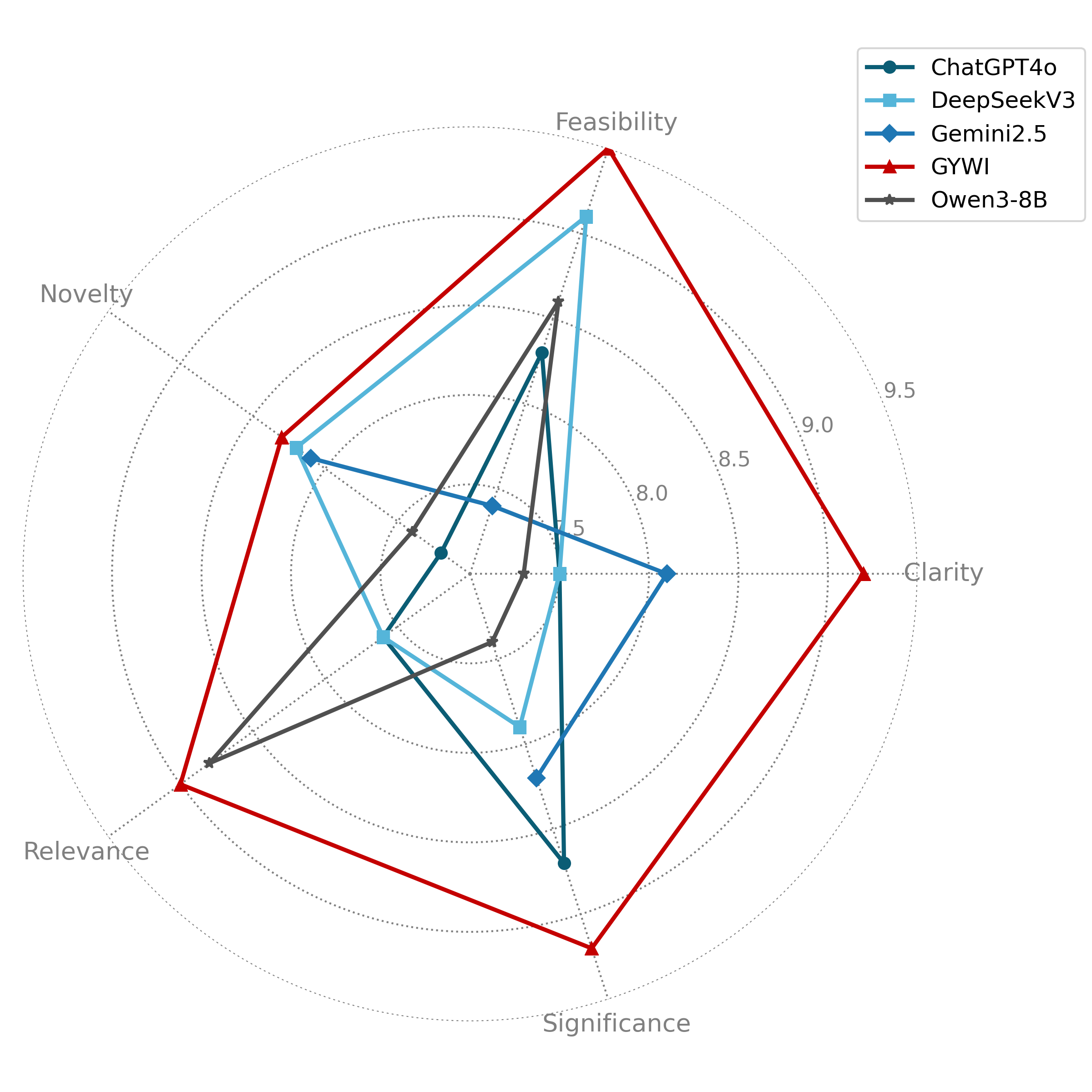}
    \caption{Human evaluation of generated ideas.}
    \label{fig:human_radar}
\end{figure}

The results of the manual evaluation show that researchers are better than LLM automatic evaluation at identifying the effectiveness of the GYWI model, amplifying the real gap between it and other models (Fig.~\ref{fig:human_radar}). The GYWI model achieved an average score of 9.04, significantly outperforming other models, especially in terms of Feasibility (9.5) and Significance (9.2).

\subsubsection{Visualising Idea Embeddings in Semantic Space}

To present the diversity of scientific ideas generated by each model in semantic space, we adopted a Sentence-BERT (\texttt{all-mpnet-base-v2}) visualization evaluation method. This method vectorises the content of the four components—\emph{Topic}, \emph{Motivation}, \emph{Novelty}, and \emph{Method} — in the research ideas generated by each model, and compresses them into a two-dimensional space using the t-SNE method for visualization. This method highlights the semantic similarity, generation stability, and diversity among the models.

Based on the aforementioned experimental setup, we randomly selected a target paper and used five models to generate 20 research ideas each for visualization evaluation.

\begin{figure}[htbp]
    \centering
    \includegraphics[width=0.8\linewidth]{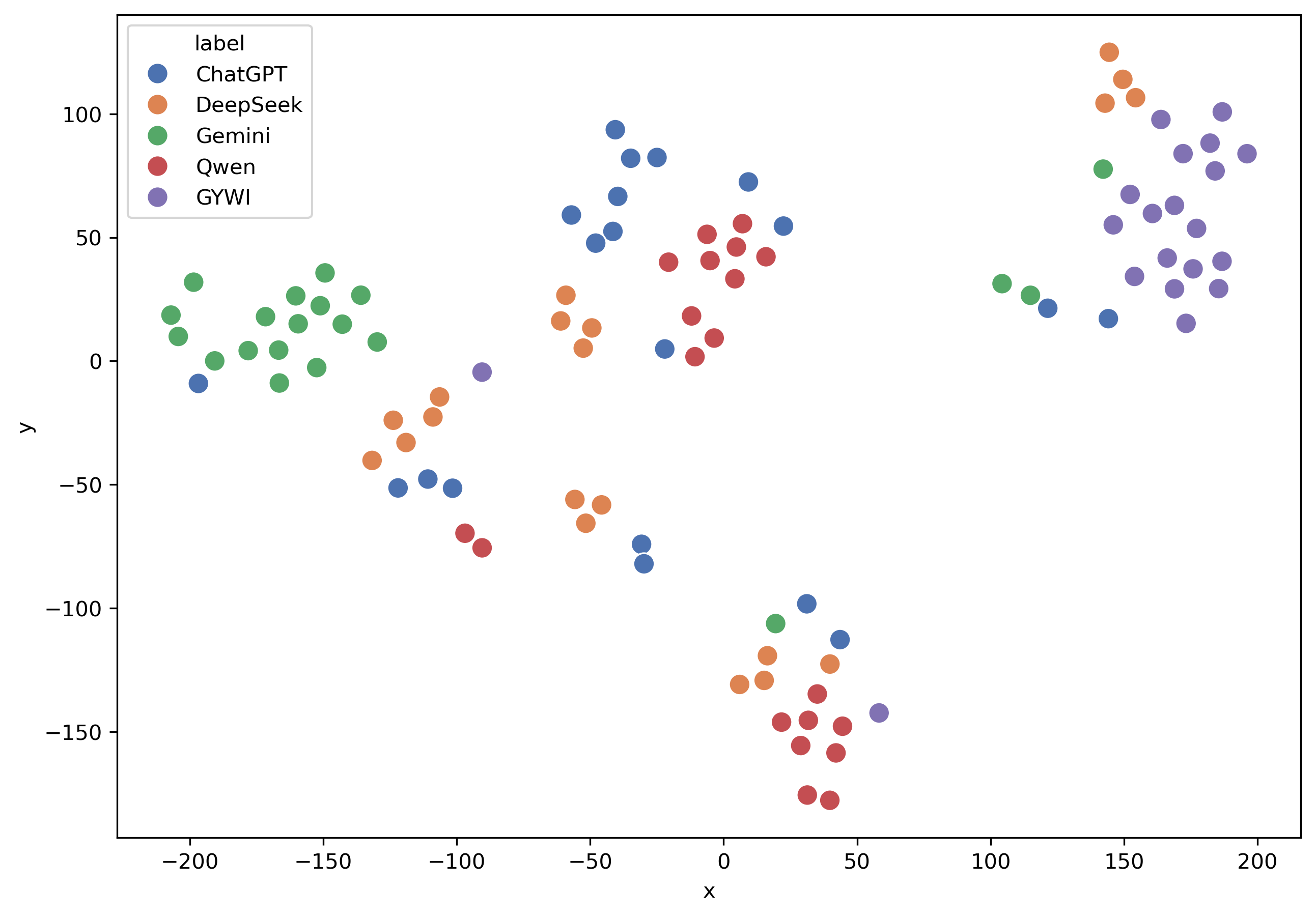}
    \caption{\textbf{Semantic Space Distribution of Research Ideas}\\
    \small Each point represents a research idea, illustrating semantic clustering and diversity between models.}
    \label{fig:semantic_space_distribution}
\end{figure}

The visualization results (Fig.~\ref{fig:semantic_space_distribution}) show that GYWI's idea distribution (purple) is more concentrated and distinct from other models, indicating high semantic consistency. A small number of nodes appear in different semantic regions, reflecting diversity, likely due to the author knowledge graph and multi-hop retrieval, which also explains its high novelty and significance scores. Although built on DeepSeek-V3, GYWI displays markedly different distributions: DeepSeek-V3's outputs are sparse and loosely clustered, lacking consistency and exhibiting stylistic uncertainty, whereas GYWI achieves both higher semantic diversity and concentration, enhancing generative control and creative stability.

\subsection{Case Study: Scientific Idea Generation Based on a given Target Paper}
In this section, we will demonstrate the effectiveness of using the GYWI system to generate scientific ideas. We randomly selected a paper by Dr Huizhi Liang from our research team as the target paper and used GYWI to generate new scientific ideas, and compared them with ideas generated solely by an LLM (DeepSeek-V3).

\subsubsection{Building an Author Knowledge Graph}
\begin{figure}[htbp]
    \centering
    \includegraphics[width=0.8\linewidth]{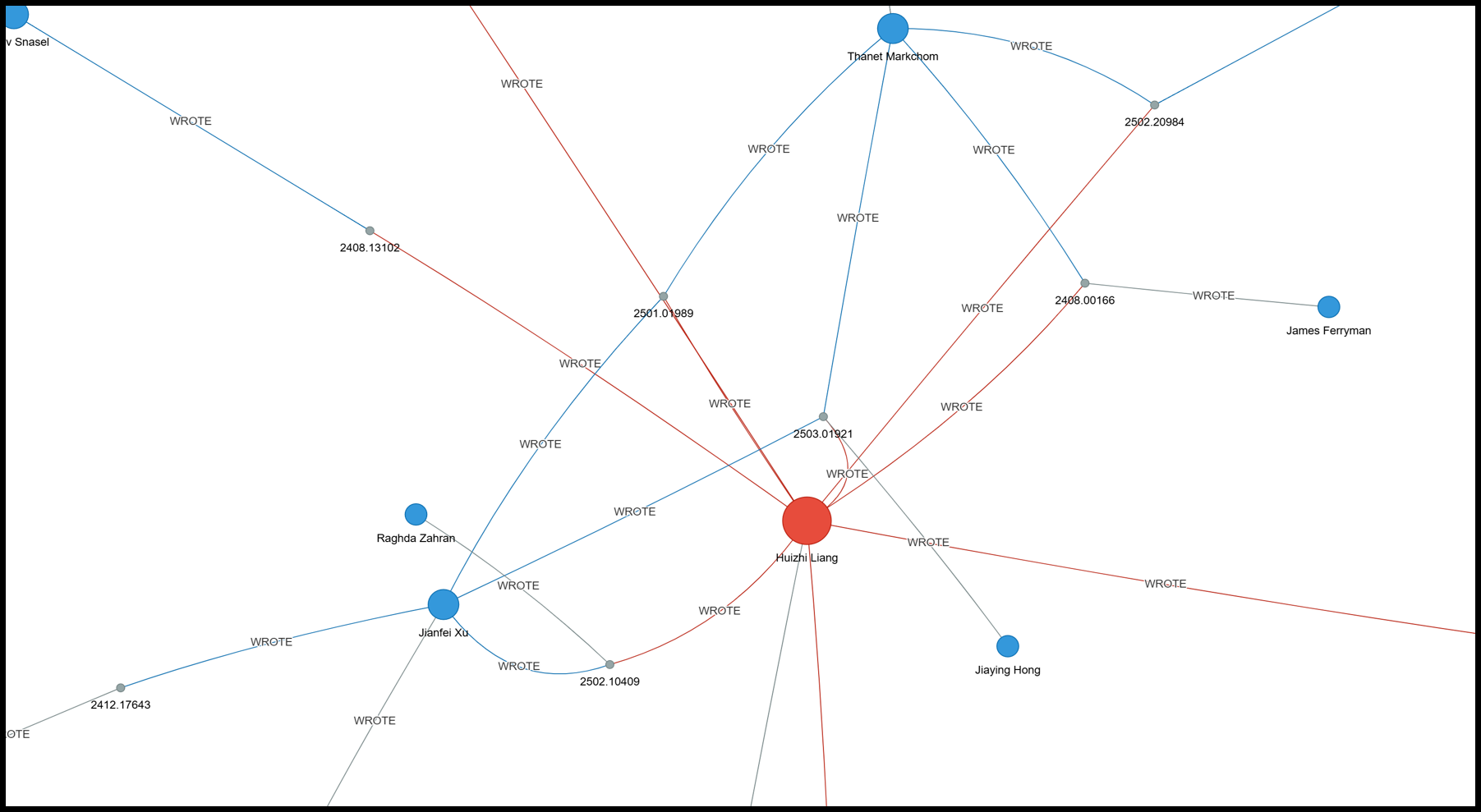}
    \caption{\textbf{Author Knowledge Graph}\\
    \small Graph of Dr Huizhi Liang's academic network, including co-authors and related papers.}
    \label{fig:graph_liang}
\end{figure}

We constructed an author knowledge graph using Yann LeCun, Geoffrey Hinton, Kaiming He, and Dr Huizhi Liang. Dr Huizhi Liang's academic network is shown in Fig.~\ref{fig:graph_liang}, which includes Dr Huizhi Liang and 85 associated authors and 47 papers. The chart shows the collaboration patterns and paper connections within the research team.

\subsubsection{Generating Research Ideas}
The retrieved information was input into the prompt optimization module, and after iteration, the system generated a new research idea. We selected the paper ``On Learnable Parameters of Optimal and Suboptimal Deep Learning Models'', which analyses the structure and operating mechanisms of deep learning models, particularly the convergence of weight distributions across architectures. Our new research explores how optimization dynamics influence universal weight patterns and whether CNT metrics can predict model performance in advance. The generation process and final results are shown in Appendix ~\ref{sec:huizhi_idea}.

\section{Conclusion}
This paper proposes GYWI, a system for generating research ideas featuring ``controllable context'' and ``traceable inspiration pathways''. It first constructs an author-centered knowledge graph, forming an external knowledge base through adjacent and random jump point sampling. Subsequently, it employs a hybrid ``depth and breadth'' retrieval approach to form the context: RAG provides fine-grained content, and GraphRAG, by organising knowledge as a graph (entity nodes, relation edges), supplies structured global connections (causal, hierarchical, co-occurrence) across the knowledge graph.Finally, a prompt optimization strategy based on reinforcement learning principles performs structured abstraction and reconstruction of retrieved content, enabling the model to better leverage relevant information. This facilitates LLM generation of research ideas with enhanced novelty, relevance, and academic feasibility. To validate effectiveness, we constructed an evaluation dataset based on arXiv papers from 2018--2023. Comprehensive assessment employed automated IMCQ objective scoring, LLM performance metrics, human evaluation, and semantic space visualization. We conducted comparative evaluation against GPT-4o, DeepSeek-V3, Qwen3-8B, and Gemini~2.5. Results demonstrate that GYWI exhibits significant advantages over mainstream baselines across multiple metrics, including novelty, reliability, and relevance.

% Subsequently, it employs a hybrid "depth and breadth" retrieval approach combining RAG and GraphRAG to generate hybrid contexts featuring both fine-grained evidence and structured global connections.

Despite performing well in multiple evaluations, GYWI retains certain limitations. 
The current framework requires a one-off construction of author graphs and retrieval indexes, 
which may pose scalability challenges as the corpus size and author network expand, 
and it currently lacks a hot-update mechanism for newly published knowledge. 
Future work may explore incremental and distributed graph construction to improve scalability, 
along with online updating mechanisms to enhance adaptability to emerging research trends and real-time data integration. 
Moreover, although preliminary experiments demonstrate GYWI’s effectiveness within specific scientific domains, 
its cross-domain generalization remains limited. 
Future research could extend GYWI by enabling retrieval and fusion across heterogeneous and domain-specific knowledge graphs, 
supporting cross-disciplinary inspiration transfer and multi-source context alignment.

\bibliographystyle{IEEEtran}
\bibliography{LOD_REFERENCES}
% \begin{thebibliography}{00}
% \bibitem{b1} G. Eason, B. Noble, and I. N. Sneddon, ``On certain integrals of Lipschitz-Hankel type involving products of Bessel functions,'' Phil. Trans. Roy. Soc. London, vol. A247, pp. 529--551, April 1955.
% \bibitem{b2} J. Clerk Maxwell, A Treatise on Electricity and Magnetism, 3rd ed., vol. 2. Oxford: Clarendon, 1892, pp.68--73.
% \bibitem{b3} I. S. Jacobs and C. P. Bean, ``Fine particles, thin films and exchange anisotropy,'' in Magnetism, vol. III, G. T. Rado and H. Suhl, Eds. New York: Academic, 1963, pp. 271--350.
% \bibitem{b4} K. Elissa, ``Title of paper if known,'' unpublished.
% \bibitem{b5} R. Nicole, ``Title of paper with only first word capitalized,'' J. Name Stand. Abbrev., in press.
% \bibitem{b6} Y. Yorozu, M. Hirano, K. Oka, and Y. Tagawa, ``Electron spectroscopy studies on magneto-optical media and plastic substrate interface,'' IEEE Transl. J. Magn. Japan, vol. 2, pp. 740--741, August 1987 [Digests 9th Annual Conf. Magnetics Japan, p. 301, 1982].
% \bibitem{b7} M. Young, The Technical Writer's Handbook. Mill Valley, CA: University Science, 1989.
% \end{thebibliography}

\appendices
\section{Materials Used in the Human Evaluation}
\label{appendix:materials}

This appendix provides the detailed materials used in the human evaluation experiment,
including:
\begin{itemize}
    \item Titles and abstracts of the target research papers.
    \item The evaluation questions provided to human annotators.
    \item The generated ideas from each evaluated model.
\end{itemize}

\subsection{Target Papers}

\paragraph{Title:} HierVL: Learning Hierarchical Video-Language Embeddings

\paragraph{Abstract:}  
Video-language embeddings are a promising avenue for injecting semantics into visual representations, but existing methods capture only short-term associations between seconds-long video clips and their accompanying text. We propose HierVL, a novel hierarchical video-language embedding that simultaneously accounts for both long-term and short-term associations. As training data, we take videos accompanied by timestamped text descriptions of human actions, together with a high-level text summary of the activity throughout the long video (as are available in Ego4D). We introduce a hierarchical contrastive training objective that encourages text-visual alignment at both the clip level and video level. While the clip-level constraints use the step-by-step descriptions to capture what is happening in that instant, the video-level constraints use the summary text to capture why it is happening, i.e., the broader context for the activity and the intent of the actor. Our hierarchical scheme yields a clip representation that outperforms its single-level counterpart as well as a long-term video representation that achieves state-of-the-art results on tasks requiring long-term video modeling. HierVL successfully transfers to multiple challenging downstream tasks (in EPIC-KITCHENS-100, Charades-Ego, HowTo100M) in both zero-shot and fine-tuned settings.

\section{Evaluation Questions}
You are provided with five research ideas generated from the title and abstract of a target research paper. Each idea is produced by a different model. Please carefully read all five ideas, compare them, and evaluate each one independently according to the following five dimensions:

\begin{enumerate}
    \item \textbf{Novelty} --- The extent to which the idea represents a creative or unexplored research direction.
    \item \textbf{Feasibility} --- The likelihood that the idea can be realistically implemented.
    \item \textbf{Clarity} --- How clearly and concisely the idea is expressed.
    \item \textbf{Relevance} --- The degree to which the idea is highly relevant to the original paper's theme.
    \item \textbf{Significance} --- The potential research value and importance of the idea.
\end{enumerate}

\noindent\textbf{Scoring scale:} For each dimension, assign a score from 1 (lowest) to 10 (highest), where 10 indicates the best possible performance in that dimension. Below are the five research ideas generated by different models, labeled as A, B, C, D, and E.

\subsection{Generated Ideas}

\paragraph{GYWI: Can hierarchical video-language embeddings be enhanced by integrating LLM-generated synthetic summaries and intermediate temporal scales to improve long-term intent modeling?}

\textbf{Motivation:}  
The target paper's hierarchical approach excels at aligning clip-level (“what”) and video-level (“why”) semantics but lacks theoretical grounding in cognitive plausibility and misses opportunities to leverage synthetic data. Additionally, the rigid two-level hierarchy may not capture intermediate temporal abstractions (e.g., “scene-level” semantics). This gap limits nuanced understanding of activities like “preparing a meal,” where mid-level goals (e.g., “chopping vegetables”) bridge instant actions and overarching intent.

\textbf{Novelty:}  
(1) \textit{Synthetic Hierarchy Injection}: Integrates LLM-generated intermediate summaries to densify supervision, inspired by synthetic narrations in related work.  
(2) \textit{Temporal Abstraction Gradients}: Introduces learnable granularity weights to dynamically adjust hierarchy depth based on video complexity.  
(3) \textit{Cognitive Validation}: Adds theory-driven metrics (e.g., intent-action consistency scores) to assess alignment with human hierarchical reasoning.

\textbf{Method:}  
Extend HierVL's contrastive loss with:  
(a) A third alignment tier for LLM-generated scene descriptions.  
(b) Adaptive attention mechanisms to weight hierarchy levels per video segment.  
Fine-tune LLMs on Ego4D summaries to generate plausible intermediate descriptions. Use Gumbel-Softmax to sample optimal hierarchy depth during training. Benchmark on EPIC-KITCHENS-100 with action-chain coherence metrics and zero-shot transfer to VidSitu.

\textbf{Difference from Original Work:}  
(1) Adds synthetic and intermediate tiers beyond human-annotated summaries.  
(2) Introduces dynamic hierarchy depth instead of fixed two levels.  
(3) Explicitly models cognitive plausibility.  
(4) Leverages LLM-augmented training rather than purely human annotations.

\textbf{Feasibility:}  
Requires LLM API access and approximately 100 GPU hours for hierarchical contrastive training. Main risks include LLM hallucinations, mitigated by prompt engineering and human verification. Alternatives include using smaller open-source LLMs. Estimated timeline: 6--12 months.

\paragraph{Gemini2.5: Goal-Driven Hierarchical Video-Language Understanding with Predictive Intent Modeling}

\textbf{Motivation:}  
The original HierVL paper captures both short-term “what” and long-term “why” associations but focuses on descriptive summaries. In real-world scenarios, understanding requires inferring underlying goals and predicting future actions or sub-tasks. This is critical for intelligent assistance, human-robot collaboration, and instructional video analysis.

\textbf{Novelty:}  
(1) \textit{Explicit Goal-Action Hierarchy Learning}: Learn a goal-action graph showing how actions contribute to sub-goals and high-level goals.  
(2) \textit{Predictive Objective for Goal Fulfillment and Next Actions}: Train to predict future actions or sub-goals given current state and inferred goals.  
(3) \textit{Goal-Conditioned Embeddings}: Represent actions differently depending on the inferred goal.

\textbf{Method:}  
Use a multi-stream hierarchical encoder similar to HierVL with an additional goal inference module (Transformer or GNN). Maintain clip-text and video-summary alignment while adding:  
(a) Predictive Goal Path Loss: Predict next action/sub-goal.  
(b) Goal Fulfillment Likelihood Loss: Estimate the probability of achieving a high-level goal.

\textbf{Difference from Original Work:}  
HierVL learns descriptive hierarchies; this idea learns predictive, inferential hierarchies centered on goals, with goal-conditioned embeddings and predictive objectives.

\textbf{Feasibility:}  
Datasets like Ego4D and EPIC-KITCHENS-100 can support this with additional annotations or weak supervision. Core components are established in current research. Computational needs are comparable to HierVL. Evaluation can include goal prediction accuracy, next-action prediction, and goal-conditioned downstream performance.

\paragraph{Qwen3-8B: Temporal-Aware Hierarchical Video-Language Embeddings for Cross-Modal Retrieval and Reasoning}

\textbf{Motivation:}  
HierVL focuses on alignment and representation but not on reasoning or retrieval. Enhancing with temporal reasoning would improve understanding of event sequences and enable complex cross-modal retrieval.

\textbf{Novelty:}  
(1) Integrate temporal reasoning modules to model causal relationships between events.  
(2) Design a cross-modal retrieval framework leveraging both clip and video-level embeddings.  
(3) Introduce a training objective encouraging sequence and intent understanding.

\textbf{Method:}  
Add a temporal reasoning module (Transformer or TCN) into HierVL's structure. Design retrieval to match queries with temporal constraints (e.g., “after the person picked up the object”).

\textbf{Difference from Original Work:}  
Adds reasoning and retrieval capabilities beyond alignment and representation.

\textbf{Feasibility:}  
Feasible with existing architectures and datasets such as Ego4D and HowTo100M.

\paragraph{DeepSeek-V3: HierVL-MultiModal: Hierarchical Video-Language-Audio Embeddings for Multimodal Context Understanding}

\textbf{Motivation:}  
HierVL models video-text relationships but ignores audio, which provides complementary cues.

\textbf{Novelty:}  
(1) Tri-modal hierarchical alignment including audio.  
(2) Cross-modal attention for dynamic fusion.  
(3) Use audio as a long-term context signal.

\textbf{Method:}  
Pretrain with clip-level and video-level audio alignment, including ASR transcripts or sound-class labels. Fuse audio, text, and visuals with cross-modal attention.

\textbf{Difference from Original Work:}  
Adds a new modality and fusion mechanisms.

\textbf{Feasibility:}  
Uses standard audio processing (e.g., Librosa) and multimodal transformers. Suitable datasets include AudioSet and EPIC-SOUNDS.

\paragraph{GPT-4o: Multi-Modal Temporal Dynamics in Hierarchical Video-Language Embeddings}

\textbf{Motivation:}  
Incorporate audio and scene dynamics to capture nuanced temporal aspects.

\textbf{Novelty:}  
(1) Audio-visual embeddings within a hierarchical framework.  
(2) Temporal attention to dynamically weigh modalities.  
(3) Cross-modal temporal consistency module.

\textbf{Method:}  
Extend contrastive objectives to include audio at clip and video levels. Integrate temporal attention for salient feature focus.

\textbf{Difference from Original Work:}  
Adds multi-modal integration with an emphasis on temporal consistency.

\textbf{Feasibility:}  
Feasible with existing deep learning tools and pretrained audio encoders like VGGish.

\section{Scientific Idea Generation Example Based on Huizhi Liang's Paper}
\label{sec:huizhi_idea}
We randomly selected the ID and title:

\begin{lstlisting}[breaklines=true]
PAPER_ID = "2408.11720"
TITLE = "On Learnable Parameters of Optimal and Suboptimal Deep Learning Models"
\end{lstlisting}

\noindent\textbf{Abstract:}
\begin{quote}
We scrutinize the structural and operational aspects of deep learning models, particularly focusing on the nuances of learnable parameters (weight) statistics, distribution, node interaction, and visualization. By establishing correlations between variance in weight patterns and overall network performance, we investigate the varying (optimal and suboptimal) performances of various deep-learning models. Our empirical analysis extends across widely recognized datasets such as MNIST, Fashion-MNIST, and CIFAR-10, and various deep learning models such as deep neural networks (DNNs), convolutional neural networks (CNNs), and vision transformer (ViT), enabling us to pinpoint characteristics of learnable parameters that correlate with successful networks. Through extensive experiments on the diverse architectures of deep learning models, we shed light on the critical factors that influence the functionality and efficiency of DNNs. Our findings reveal that successful networks, irrespective of datasets or models, are invariably similar to other successful networks in their converged weights statistics and distribution, while poor-performing networks vary in their weights. In addition, our research shows that the learnable parameters of widely varied deep learning models such as DNN, CNN, and ViT exhibit similar learning characteristics.
\end{quote}

\subsubsection{LLM-Only Baseline (DeepSeek-V3)}
\paragraph{Topic}
Understanding the Role of Weight Dynamics During Training in Optimal vs. Suboptimal Deep Learning Models

\paragraph{Motivation}
The original paper focuses on analyzing the converged weights of deep learning models to identify patterns distinguishing optimal and suboptimal networks. However, it does not explore how these weight patterns emerge during training. Understanding the dynamics of weight evolution (e.g., trajectory, stability, and transient behaviors) could provide deeper insights into why some models converge successfully while others fail, addressing a critical gap in interpretability and training optimization.

\paragraph{Novelty}
\begin{enumerate}
    \item \textbf{Temporal Analysis of Weight Dynamics}: Instead of static converged weights, study how weights evolve over training epochs, identifying key phases (e.g., rapid learning, saturation, or instability).
    \item \textbf{Early Performance Prediction}: Investigate whether early weight behavior (e.g., variance trends, gradient alignment) can predict eventual model success/failure.
    \item \textbf{Architecture-Agnostic Dynamic Signatures}: Extend findings beyond static weight statistics to dynamic patterns applicable across DNNs, CNNs, and ViTs.
\end{enumerate}

\paragraph{Method}
\begin{enumerate}
    \item \textbf{Track and Visualize Weight Trajectories}: Use tools like TensorBoard or custom logging to record weight distributions, gradients, and layer-wise interactions at regular intervals during training.
    \item \textbf{Quantify Dynamic Metrics}: Develop metrics for weight stability (e.g., epoch-to-layer variance), gradient coherence, and convergence speed, then correlate these with final test accuracy.
    \item \textbf{Controlled Experiments}: Train models with varying hyperparameters (e.g., learning rates, initialization) to isolate how dynamics influence outcomes.
\end{enumerate}

\paragraph{Difference}
The original paper studies \textbf{static} weight properties post-training, while this idea focuses on \textbf{dynamic} behavior \textbf{during} training. It also introduces predictive early signals, which the original work does not address.

\paragraph{Feasibility}
Highly feasible: Requires standard deep learning frameworks (PyTorch/TensorFlow) and tools for weight logging/visualization. Techniques like gradient tracking and trajectory analysis are well-established in related fields (e.g., optimization).

\subsubsection{GYWI-Generated Idea}
\paragraph{Topic}
How do optimization dynamics shape universal weight distribution patterns across diverse architectures, and can Complex Network Theory (CNT) metrics predict model performance a priori?

\paragraph{Motivation}
The target paper establishes that successful networks converge to similar weight distributions, but does not explain why or how optimization algorithms contribute to this phenomenon (explicit limitation). Adjacent literature shows adaptive optimizers like ADAM exhibit generalization gaps, while CNT metrics (e.g., node strength) correlate with performance. This gap presents an opportunity to unify optimization dynamics, weight distribution analysis, and CNT into a predictive framework for model evaluation.

\paragraph{Novelty}
\begin{enumerate}
    \item \textbf{Theoretical Integration}: First formal linkage between optimization trajectory (e.g., ADAM's gradient variance) and emergent CNT metrics in weight distributions (leveraging ``Conceptual Bridge'' from graphRAG).  
    \item \textbf{Architectural Agnosticism}: Extends target paper's findings to transformer-based LLMs (GPT-style) and residual networks (ResNet), testing the universality claim (``Empirical Notes'' in RAG).  
    \item \textbf{Predictive Application}: Uses CNT metrics (node strength, layer fluctuation) as early stopping criteria or architecture selection heuristics (``Innovation Seed'' in RAG).  
\end{enumerate}

\paragraph{Method}
\begin{enumerate}
    \item \textbf{Core Methodology}:  
    \begin{itemize}
        \item Track weight distribution evolution (mean/variance, t-SNE clustering) across training phases for ViT, CNN, and DNN models under SGD/ADAM/Lookahead.  
        \item Compute CNT metrics (degree centrality, modularity) from weight matrices at each epoch.  
    \end{itemize}
    \item \textbf{Integration Strategy}:  
    \begin{itemize}
        \item Apply Granger causality to test if optimizer hyperparameters (e.g., ADAM's $\beta_1$) drive CNT metric convergence.  
        \item Train logistic regression classifiers to predict final model performance from early-training CNT metrics.  
    \end{itemize}
    \item \textbf{Validation}:  
    \begin{itemize}
        \item Benchmark against target paper's datasets (CIFAR-10, MNIST) + NLP tasks (GLUE for transformers).  
        \item Compare prediction accuracy of CNT metrics vs. traditional validation loss.  
    \end{itemize}
\end{enumerate}

\paragraph{Difference}
\begin{enumerate}
    \item \textbf{Scope Expansion}: Target paper analyzes static converged weights; we study dynamic optimization paths and their CNT correlates.  
    \item \textbf{Methodological Additions}: Introduces causal analysis (Granger) and predictive modeling (logistic regression) absent in original work.  
    \item \textbf{Architectural Breadth}: Adds LLMs and ResNets, addressing the ``larger architectures'' gap noted in RAG.  
\end{enumerate}

\paragraph{Feasibility}
\begin{enumerate}
    \item \textbf{Resources}: Requires $\sim$100 GPU-hours for training diverse architectures (feasible via cloud platforms).  
    \item \textbf{Hurdles}: CNT metrics may not scale efficiently to billion-parameter models; fallback to layer-wise sampling.  
    \item \textbf{Alternatives}: If Granger causality fails, use correlation thresholds or attention map analysis for transformers.  
\end{enumerate}

\noindent\textbf{Timeline}: Medium-term (6--12 months) for initial validation; long-term for industry-scale deployment.  

\noindent\textbf{Success Metrics}:
\begin{itemize}
    \item $\geq 0.8$ AUC for performance prediction using early CNT metrics.  
    \item Significant Granger causality ($p<0.01$) between optimizer settings and CNT convergence.  
\end{itemize}

\section{Evaluation Prompt}
\label{appendix:evaluation_prompt}

\begin{lstlisting}[breaklines=true]
You are a strict research evaluation expert.

Here are 5 research ideas under the same topic: "{title}", abstract: "{abstract}".

Each idea includes: Motivation, Novelty, Method, Difference, and Feasibility.

Your task is to evaluate these 5 ideas (individually) on each of the following dimensions:
- Novelty
- Feasibility
- Clarity
- Relevance
- Significance

Score each idea on a scale from 1 to 10 (10 = best, 1 = worst) for **each** dimension.

Respond ONLY in **valid JSON format** like this (no markdown, no explanations):

{
  "Novelty": {
    "ChatGPT4o": 9,
    "Gemini2.0": 6,
    "DeepSeekV3": 7,
    "Qwen3-8B": 8,
    "GYWI": 8
  },
  ...
}

Only use exact model names from this list: {json.dumps(source_list)}

Here are the 5 ideas:
A:

B:

C:

D:

E:

\end{lstlisting}

\section{ Prompt Design Comparison}
\label{appendix:prompts}

\subsection{Original Prompt}
\begin{lstlisting}[breaklines=true]
You are an expert in scientific research ideation.

Below is the title of a published research paper:

<target_paper>
Title: {title}
</target_paper>

Your task is to carefully read and understand the main content, motivation, and contribution of this paper.

Then, based on its research theme and limitations, generate a new research idea that is:
- clearly related to the original topic
- not redundant (should not repeat the original idea)
- logically novel (provides a new perspective, application, or method)
- feasible (can be reasonably implemented with known techniques)

Please structure your response using the following XML format:
<topic>{What is the new dissertation topic}</topic>
<motivation>{What motivates this new idea, and what gap or opportunity it addresses}</motivation>
<novelty>{What is new or unique compared to the original paper? List 2-3 concrete innovations}</novelty>
<method>{A high-level description of how the new idea would be implemented. Include at least 2 core steps or modules}</method>
<difference>{How is this idea different from the original paper? Be specific}</difference>
<feasibility>{Is the idea realistic? What methods or tools would be required to realize it?}</feasibility>

Note:
- Be as specific as possible.
- Use your expert knowledge to ensure that the new idea adds real value and builds upon the original work.
- Do not repeat the content of the original paper.

Please output a new idea and a simple explanation.
\end{lstlisting}

\subsection{Optimized Prompt}
\begin{lstlisting}[breaklines=true]
You are an expert research innovator specializing in interdisciplinary synthesis. Your task is to generate novel, high-impact research ideas by combining insights from the target paper with complementary concepts from related literature.

<context>
1. Target Paper (Core Contribution):
Title: {title}
Abstract: {abstract}

2. Conceptual Adjacencies (graphRAG Synthesis):
{graph_summary}

3. Foundational Elements (RAG Chunks):
{chunks}
</context>

Generate a research extension that:
1. Creates meaningful synergy between the target paper's approach and at least one adjacent concept
2. Addresses either: (a) an explicit limitation in the original work, or (b) an untapped application domain
3. Demonstrates clear academic value through either theoretical advancement or practical applicability

Quality Criteria:
- Novelty: Must introduce at least one genuinely new element (method, application, or theoretical integration)
- Feasibility: Should leverage established methods while pushing boundaries
- Impact: Potential to open new research directions or solve important problems
- Specificity: Must include concrete methodological details and validation approaches

Additional Requirements:
- Flag any conceptual risks or open challenges
- Suggest 1-2 metrics for evaluating success
- Indicate expected development timeline (short/medium/long-term)

Please structure your response using the following XML format:
<topic>{{Precisely formulated research question}}</topic>
<motivation>{{Clear gap analysis with references to specific context elements (max 150 words)}}</motivation>
<novelty>{{2-3 specific innovations with explicit ties to graphRAG/RAG concepts}}</novelty>
<method>{{Implementation blueprint: (1) Core methodology, (2) Integration strategy, (3) Validation approach}}</method>
<difference>{{Point-by-point comparison with target paper's scope and methods}}</difference>
<feasibility>{{Realistic assessment including: (1) Required resources, (2) Technical hurdles, (3) Alternative pathways}}</feasibility>
\end{lstlisting}

\subsection{Intermediate Optimization Prompt}
\begin{lstlisting}[breaklines=true]
You are a research summarization assistant. Your goal is to help a synthesis model use contextual documents more effectively.

<task>
Given:
1. A GraphRAG conceptual summary (already compressed and high-level)
2. Several RAG semantic chunks (noisy but informative, contains the paper itself, adjacent and random information. Random information is necessary in the hope of getting some innovative inspiration.)
3. A target paper (title + abstract)

Please:
- Keep the GraphRAG summary mostly intact, but rephrase or re-organize it for clarity if needed
- Carefully select and compress the most relevant RAG chunks
- Do NOT merge GraphRAG and RAG into one block; treat them as complementary layers of context
- Output a structured 2-section summary:
   [GraphRAG Context]
   [RAG Context]

<target_paper>
Title: {title}
Abstract: {abstract}
</target_paper>

<GraphRAG_Summary>
{graph_summary}
</GraphRAG_Summary>

<RAG_Chunks>
{chunk_text}
</RAG_Chunks>

Return ONLY the combined two-part summary in plain text with clear section labels.
\end{lstlisting}

\section{Example of IMCQ Question}
\label{appendix:imcq_example}

\begin{lstlisting}[breaklines=true]
{
  "index": 0,
  "title": "Robust Pre-Training by Adversarial Contrastive Learning",
  "id": 2010.13337,
  "abstract": "Recent work has shown that, when integrated with adversarial training, selfsupervised pre-training can lead to state-of-the-art robustness",
  "question": {
    "a": "Despite the remarkable advancements in self-supervised pre-training combined with adversarial training, existing approaches often fail to achieve a balance between robustness and generalizability. Previous studies primarily focus on either achieving robustness or optimizing contrastive learning independently, leaving a crucial gap in their joint optimization. This paper seeks to address this limitation by exploring how to effectively integrate adversarial contrastive learning to enhance overall robustness during the pre-training phase, ensuring improved performance across diverse scenarios.",
    "b": "Current self-supervised learning (SSL) in medical image analysis is mostly confined to single modalities, limiting its universality across different modalities. Joint SSL, which combines all modality data, faces challenges like modal data collision and high costs for integrating new knowledge. MedCoSS proposes a sequential pre-training approach to mitigate these issues, enhancing the model's generalization and scalability across various medical modalities.",
    "c": "Adversarial learning has shown success in various domains but applying it to pre-training LLMs is impractical due to high data and computational costs. ICL is effective but sensitive to prompt changes. adv-ICL addresses these issues by optimizing prompts in an adversarial manner without updating model parameters, making it computationally efficient and effective in low-resource settings.",
    "d": "Deep neural networks (DNNs) are vulnerable to adversarial attacks, hindering their reliable deployment. Existing defenses often involve cumbersome training processes or domain-specific heuristics, lacking efficiency and adaptability. This research addresses the open challenge of reprogramming well-trained models for robustness without parameter modification, offering an efficient and innovative approach to adversarial defenses."
  },
  "answer_option": "a"
}
\end{lstlisting}

\end{document}